# Separation of water and fat signal in whole-body gradient echo scans using convolutional neural networks

**Authors:** Jonathan Andersson[1]*, Håkan Ahlström[1,2], and Joel Kullberg[1,2]

[1]Section of Radiology, Department of Surgical Sciences, Uppsala University, Uppsala, Sweden.

[2]Antaros Medical, Mölndal, Sweden.

*Correspondence to: Jonathan Andersson, MRT, Entrance 24, Uppsala University Hospital, SE-751 85 Uppsala, Sweden. Email: jonathan.andersson@surgsci.uu.se

# Abstract

**Purpose:** To perform and evaluate water–fat signal separation of whole-body gradient echo scans using convolutional neural networks.

**Methods:** Whole-body gradient echo scans of 240 subjects, each consisting of 5 bipolar echoes, were used. Reference fat fraction maps were created using a conventional method. Convolutional neural networks, more specifically 2D U-nets, were trained using 5-fold cross-validation with 1 or several echoes as input, using the squared difference between the output and the reference fat fraction maps as the loss function. The outputs of the networks were assessed by the loss function, measured liver fat fractions, and visually. Training was performed using a graphics processing unit (GPU). Inference was performed using the GPU as well as a central processing unit (CPU).

**Results:** The loss curves indicated convergence, and the final loss of the validation data decreased when using more echoes as input. The liver fat fractions could be estimated using only 1 echo, but results were improved by use of more echoes. Visual assessment found the quality of the outputs of the networks to be similar to the reference even when using only 1 echo, with slight improvements when using more echoes. Training a network took at most 28.6 h. Inference time of a whole-body scan took at most 3.7 s using the GPU and 5.8 min using the CPU.

**Conclusion:** It is possible to perform water–fat signal separation of whole-body gradient echo scans using convolutional neural networks. Separation was possible using only 1 echo, although using more echoes improved the results.

# 1 Introduction

The vast majority of the signal in $^1$H MRI of humans without implants originate from either water or fat molecules. It is often of interest, both in clinical practice and in research studies, to separate the water and the fat signal from each other. For certain types of scans, this can be performed in post-processing by using the property of chemical shift, which was first proposed by Dixon in 1984 (1).

The methods used for water–fat signal separation have since been refined. The most important addition has been taking the amplitude of the static magnetic field ($B_0$) inhomogeneity into account, without which the signal separation will be incomplete (2). The inclusion of the effective transverse relaxation rate ($R_2^*$) and a multi-peak fat spectrum results in an even more complete signal separation (3).

After signal separation, it is possible to calculate the percentage of the total signal originating from fat, the so-called fat fraction, which is a useful quantitative measurement. As an example, the fat fraction of the liver can be used to evaluate hepatic steatosis, thereby avoiding biopsies (4).

To perform the signal separation, at least 2 echoes are needed or else the problem is underdetermined. However, a few methods have been developed to perform the signal separation using a single echo by making assumptions of the composition of the voxels (5, 6). The assumptions can lead to severe errors where they are not valid, which is probably the main reason why these methods are not commonly used.

The signal separation can be performed using either gradient or spin echoes. When using a normal Cartesian k-space trajectory, the gradient echo sequences will produce echoes of 2 different polarities. Using only echoes of 1 polarity avoids the problems associated with signal separation of bipolar echoes. Echoes of opposing polarities will have the water–fat signal shift in opposite directions, differences in the signal strength because of frequency dependent coil sensitivities, and spatial distortions in opposite directions caused by field inhomogeneities. Finally, and often most importantly, polarity dependent phase errors induced by eddy currents have to be considered when 3 or more bipolar echoes are used (7). However, even when taking the polarity dependent amplitude and phase errors into account, there may be a fat fraction dependent bias when using bipolar errors that is greater than when using monopolar echoes (7).

Recently, a class of machine learning algorithms called artificial neural networks, often shortened to neural networks or even just networks, have become extremely popular, especially within image processing. This is because of their often excellent performance compared to other machine learning algorithms. Today, virtually all neural networks contain multiple hidden layers and therefore fall under the category deep learning. Within image processing, so-called convolutional neural networks are commonly used (8). Convolutional neural networks typically take an image as input (e.g., an image of a person), and the output is typically a class (e.g., the gender of the person), a number (e.g., the age of the person), a segmentation, or a translated image (e.g., colorizing a black and white image). One widely used type of convolutional neural networks are the so-called U-nets, which were originally designed for image segmentation (9). U-nets have since been used for multiple different image segmentation tasks (e.g., automated segmentation of abdominal adipose tissue depots in water–fat separated MR images) (10). U-net based architectures have also been used in image-to-image translation tasks (11).

Neural networks have been used within MR image reconstruction to transform data from k-space to image-space (12, 13), calculate parametric maps in MRI fingerprinting (14), and recent conference abstracts show promise in water–fat signal separation (15–18). The authors of one of the abstracts (16) have since published a manuscript on the same topic (19).

In this article, a method using neural networks, specifically U-nets, for separation of water–fat signal in whole-body gradient echo images is presented and evaluated. This task is a type of image-to-image translation. The method builds on a previous conference abstract (15). Separation is performed using both a single echo as well as multiple echoes.

## 2 Methods

2.1 Source data

Whole-body imaging data from the POEM study (20), where all subjects are 50 years old, was used. In this article, a total of 240 scans, each of a different subject, were included after removal of scans of poor quality. Poor quality included excessive motion artifacts and errors in the scanning protocol, minor metal artifacts were accepted. Approval of the POEM study was obtained from the regional ethics committee, and each participant gave their written informed consent.

The images were all acquired on a 1.5T clinical scanner (Achieva, Philips Healthcare, Best, The Netherlands). A 3D spoiled gradient echo sequence was used. A total of 5 bipolar echoes were collected. The following parameters were used: voxel size = 2.07 × 2.07 × 8 mm$^3$ (sagittal × coronal × axial), TE$_1$ = 1.37 ms, ΔTE = 0.95 ms, TR in range: 6.65–7.17 ms, and flip angle = 3°. The images were collected with continuously moving bed imaging (21), resulting in several subvolumes. The whole-body images were of size 256 × 184 × 252 voxels.

## 2.2 Reference method

The 3 odd-numbered echoes were used to create reference signal separations using the previously described analytical graph-cut method (22), producing water and fat images, as well as fat fraction, $R_2^*$, and field maps. One subvolume was processed at a time. Because of noise and model imperfections, the fat fraction maps could contain values lower than 0% or higher than 100%, these values were set to 0% and 100%, respectively.

## 2.3 Neural networks

Modified versions of the U-net (9) were used in this article. They were trained with axial slices of different sets of echoes as input and the corresponding fat fraction maps as the desired output.

The U-net is described in detail in the original article (9), but will be described in brief here. An input image will go through convolutions, producing multiple features, after which the features are downsampled to a lower resolution. This process is repeated a few times. After this, the resulting features undergo some more convolutions, and the resulting features are then upsampled, after which a concatenation with the previous features of the same resolution is performed using so-called skip connections. This process is repeated until features are produced of the same resolution as the original input image. Finally, some more convolutions are performed to produce the output. A visual representation of a modified U-net used in this article can be seen in Figure 1.

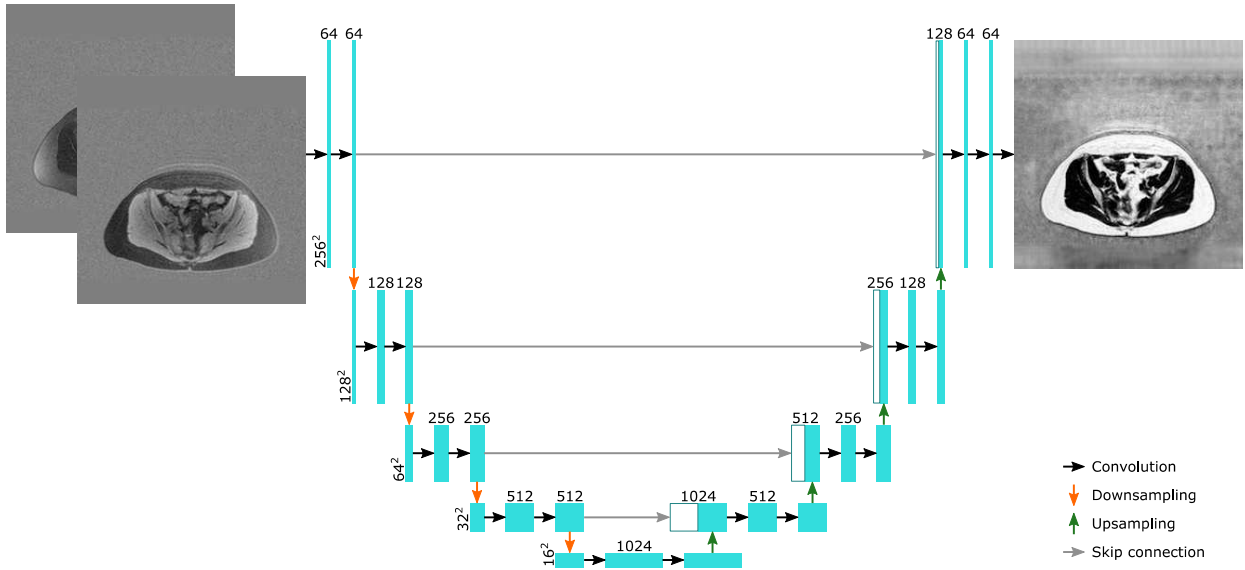

**Figure 1.** A visual representation of one of the networks used in this manuscript, with an axial slice containing the real and the imaginary parts of the first echo as input and the corresponding fat fraction map as output. The cyan boxes represent feature maps. The white boxes represent feature maps that have been transferred by the skip connections. The horizontal numbers represent the number of features in a layer and the vertical numbers represent the number of elements per feature of the layer.

The networks used in this article were implemented in TensorFlow 1.8.0 (23), and were based on an implementation by Akeret et al. (24) Differences compared to the original implementation by Ronneberger et al. (9) will be described, but the implementations are otherwise identical.

As input to the networks 1 2D axial slice with 2 channels for each echo, 1 for the real and 1 for the imaginary component, was used. The input was scaled to be within the range −1 to 1, with 0 representing no signal and zero-padded to be of size 256 × 256.

The networks were trained using different configurations of the available echoes as input. Networks were either trained with echoes of both polarities or only echoes of 1 polarity (i.e., only odd or only even numbered echoes). For the 3 different sets of echoes used, all possible configurations using the first available echo and different numbers of consecutive echoes were used. Using echoes of both polarities explores if and how much this might improve the resulting signal separation.

The original implementation of the U-net would return an output that was smaller than the input (i.e., the result would be cropped) because of the convolutions. This was not desirable for the current problem and was rectified by performing reflective paddings inside the networks after the convolutions, which made the output the same size as the input.

The implemented networks had 1 feature map as output. As loss function, the voxelwise squared difference between the reference fat fraction, scaled to the range 0 to 1, and the output was used. This implies that the networks were trained to calculate fat fraction maps. Background voxels were excluded when calculating the loss function because they contain only noise and could potentially interfere with the training of the networks. Background was defined slicewise using Otsu's threshold (25) on the sum of the reference water and fat images. Slices containing only background were semi-automatically identified and excluded.

The networks were trained using a 5-fold cross-validation, split at the subject level (i.e., 80% of all subjects were used for training and 20% for validation), and this was repeated 5 times so that all data was used in the validation. The split was randomized. All slices of the training sets were used to perform training 1 at a time in a random order. One pass over all these slices is known as one epoch. The Adam optimizer (26) was used to train the networks. The following parameters were used: initial learning rate: 0.001, $\beta_1$ = 0.9, $\beta_2$ = 0.999, $\varepsilon$ = 1e−8. The learning rate was decayed by a factor 0.8727 after each epoch.

The networks were trained for 16 epochs. After each epoch, every second slice that was used for validation and every eighth slice that was used for training were run through the network to calculate loss curves.

2.4 Water–fat signal images

Water–fat signal images can be created from the resultant fat fraction maps of the trained networks. The signal of the echoes of bipolar spoiled gradient echo sequences can be described as

$$S_n = (W + a_n F)e^{i\omega_0 + (i\omega - R_2^*)t_n + (-1)^n \theta} \qquad [1]$$

where $S_n$ is the signal at echo time $t_n$, $W$ and $F$ are the magnitudes of the water and the fat signals, respectively, $\omega_0$ describe the initial phase of the signal, $\omega$ the off-resonance shift, $R_2^*$ is the effective transverse relaxation rate, and $\theta$ is a complex value describing the polarity-dependent amplitude and phase of the signal. $a_n$ is

$$a_n = \sum_{m=1}^{M} \alpha_m e^{i \gamma B_0 \delta_m t_n} \qquad [2]$$

where $α_m$ are the relative magnitudes of the $M$ different fat peaks and $δ_m$ are their corresponding chemical shifts relative to water. Values were adapted from Hamilton et al. (27). $B_0$ is the amplitude of the static magnetic field and γ the gyromagnetic ratio of $^1H$.

If the fat fraction (*FF*) is defined as $F/(W + F)$, and $R_2^*$ and the real part of θ assumed to be zero, it is possible to calculate *W* and *F* as

$$W = \left|\frac{S_n(1-FF)}{1+FF(a_n-1)}\right| \quad [3]$$

and

$$F = \left|\frac{S_n*FF}{1+FF(a_n-1)}\right| \quad [4]$$

In case multiple echoes are used, *W* and *F* can be calculated as the average for the different echoes to improve the SNR.

## 2.5 Hardware

All networks were trained using a graphics processing unit (GPU) of type GeForce GTX 1080 Ti. Furthermore, inference was performed using both the GPU as well as a central processing unit (CPU) of type Intel Xeon W-2102.

## 2.6 Evaluation

The final values of the loss function for the subjects used for validation is a measure of how well the networks performed. In addition to this, the quality of the outputs of the networks were assessed by measured liver fat fractions and visual inspection.

To perform the measures of the liver fat fraction, the livers were manually segmented. The fat fractions of the livers were calculated as the median of all voxels that were segmented. The performances of the different networks were evaluated by calculating the mean absolute error. Furthermore, it was evaluated how well the networks classified the fat fractions of the livers as normal or abnormal/fatty, using the commonly used cut-off value of 5.56% (4).

Visual evaluation consisted of searching for errors in the images inferred by the networks, as well as finding qualitative differences compared to the reference images.

The time taken to train the networks with different number of echoes were measured, including the time taken to produce the loss curves. Inference time was also measured.

## 3 Results

All results will refer to the output of the fully trained neural networks with validation data as input unless otherwise stated.

3.1 Loss function

In Figure 2, the loss curves for the networks using echoes of both polarities are shown. It can be seen that after a few epochs the curves for the validation data flatten out even though the curves for the training data continue to decrease. This indicates that no overfitting has taken place, and the output of the networks converged for the validation data.

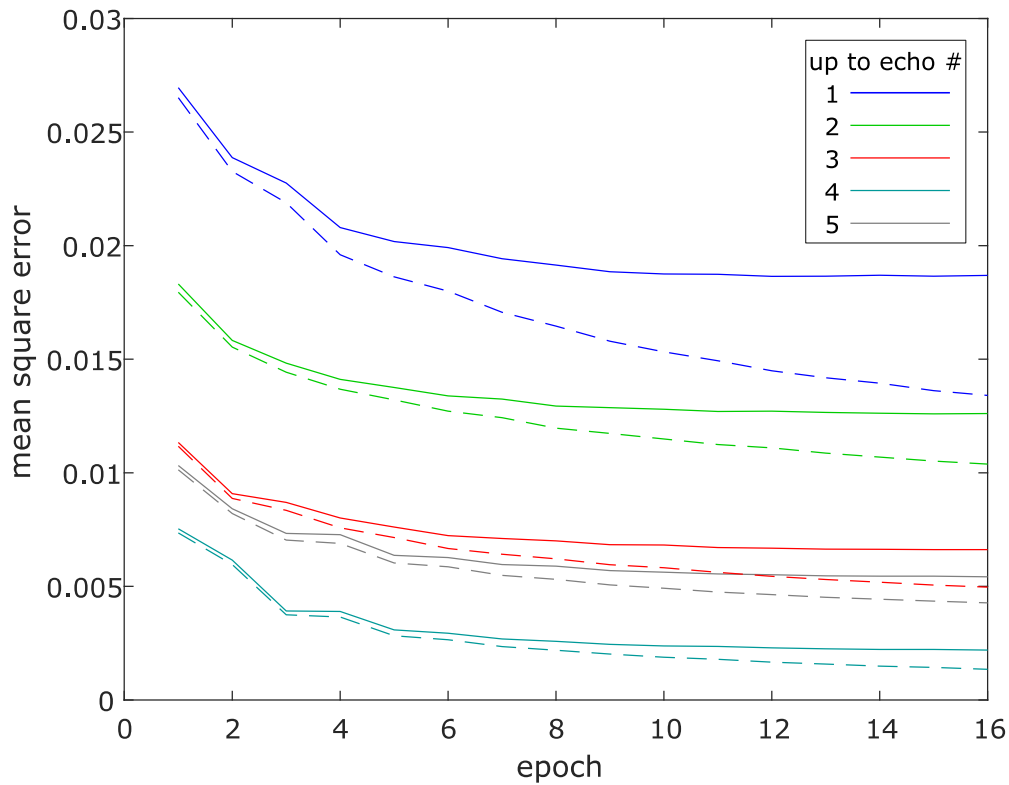

**Figure 2.** Curves showing loss per foreground voxel for the networks using echoes of both polarities. Dashed lines are used for training data, solid lines for validation data.

In Table 1, the final losses per foreground voxel of the validation data for all the different configurations of echoes used are shown.

**Table 1.** Final losses per foreground voxel of the validation data.

| Up to echo no. | 1 | 2 | 3 | 4 | 5 |
|---|---|---|---|---|---|
| All echoes | 0.0187 | 0.0126 | 0.0066 | 0.0054 | 0.0022 |
| Odd echoes | 0.0187 | – | 0.0069 | – | 0.0022 |
| Even echoes | – | 0.0165 | – | 0.0113 | – |

3.2 Liver fat content

When evaluating the networks' ability to calculate the liver fat fractions, it was noticed that the scans of 2 subject were faulty, probably because of errors in the scanning protocol, and they are not included in the results regarding the livers, leaving 238 subjects. In Figure 3, the liver fat fractions estimated by the neural networks using echoes of both polarities are plotted against the reference fat fraction. It can be seen that even when using only 1 echo, it is possible to estimate the liver fat fractions. The estimates improve with more echoes and are almost identical to the reference when using all 5 echoes.

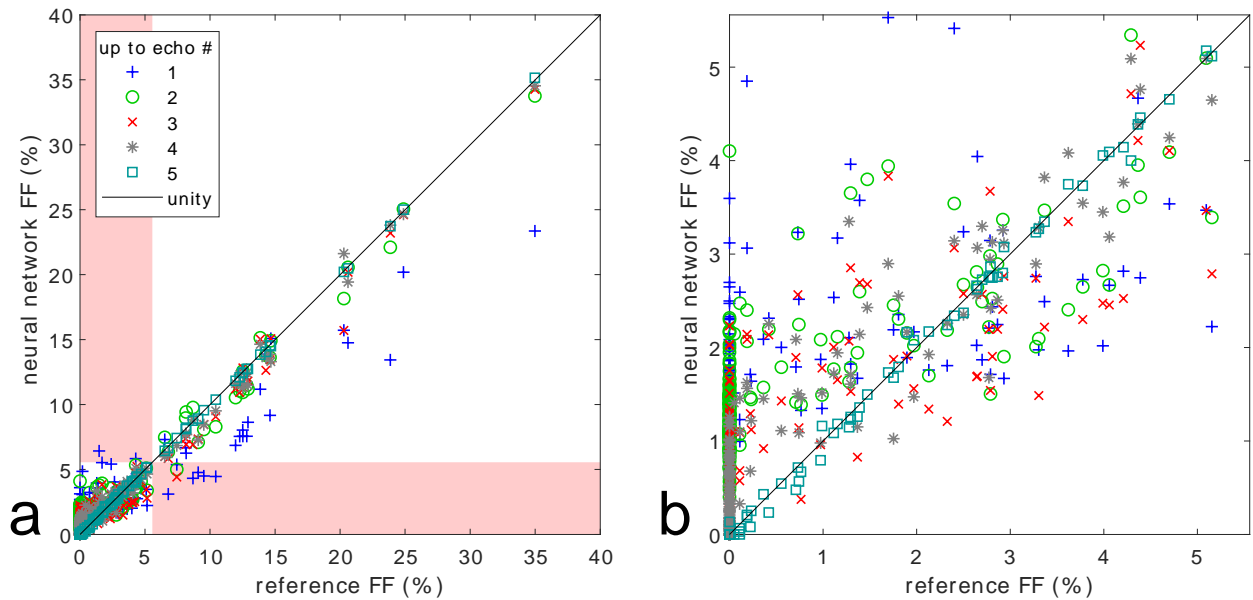

**Figure 3.** Liver fat fractions estimated by the neural networks using echoes of both polarities plotted against the reference fat fraction. **a**, All data points shown. In case a normal liver was misclassified as fatty or vice versa, the corresponding data point is found in the red shaded area. **b**, Zoom-in on the area

with the cases were the liver fat fractions were correctly classified as normal, where most cases can be found.

According to the reference fat fraction, the liver fat fraction was normal for 214 of the subjects and abnormal for the remaining 24. Table 2 shows the mean absolute errors of the liver fat fraction calculated using the neural networks, and Table 3 shows the number of misclassified livers. Both tables make it clear that by using more echoes the results improved.

**Table 2.** Mean absolute errors of the liver fat fraction calculated using the neural networks.

| Up to echo no. | 1 | 2 | 3 | 4 | 5 |
|---|---|---|---|---|---|
| All echoes (%) | 1.71 | 1.18 | 0.69 | 0.39 | 0.03 |
| Odd echoes (%) | 1.71 | – | 0.93 | – | 0.04 |
| Even echoes (%) | – | 1.52 | – | 1.22 | – |

**Table 3.** Numbers of livers misclassified as normal/fatty.

| Up to echo no. | 1 | 2 | 3 | 4 | 5 |
|---|---|---|---|---|---|
| All echoes | 6/2 | 1/0 | 1/0 | 0/0 | 0/0 |
| Odd echoes | 6/2 | – | 5/1 | – | 0/0 |
| Even echoes | – | 5/1 | – | 0/0 | – |

According to the reference fat fraction, 214 were normal and 24 were fatty.

According to Reeder et al. (28), the accuracy (bias) and precision (SD) of a quantitative fat content biomarker must be far smaller than 5–6% to provide reliable diagnosis. In Table 2, it can be seen that for all networks the accuracy (bias) of the estimated liver fat fraction, compared to the reference, filled this criteria. It was not possible to evaluate whether the precision (SD) of the estimated liver fat fractions filled the criteria because each individual was only scanned once, and it was therefore not possible to calculate the precision (SD).

3.3 Visual inspection

In Figure 4, axial fat fraction maps of the abdomen (including the liver) of a subject with a fatty liver is shown. Both the fat fraction maps inferred by the networks using echoes of both polarities as input and the reference fat fraction map are shown. Two improvements are noticeable when increasing the number of echoes used. First, the images get crisper, and second, the fat fraction of the liver gets closer to the reference. When using all 5 echoes, the inferred image is almost identical to the reference. The images are representative, with the exception of the high liver fat content.

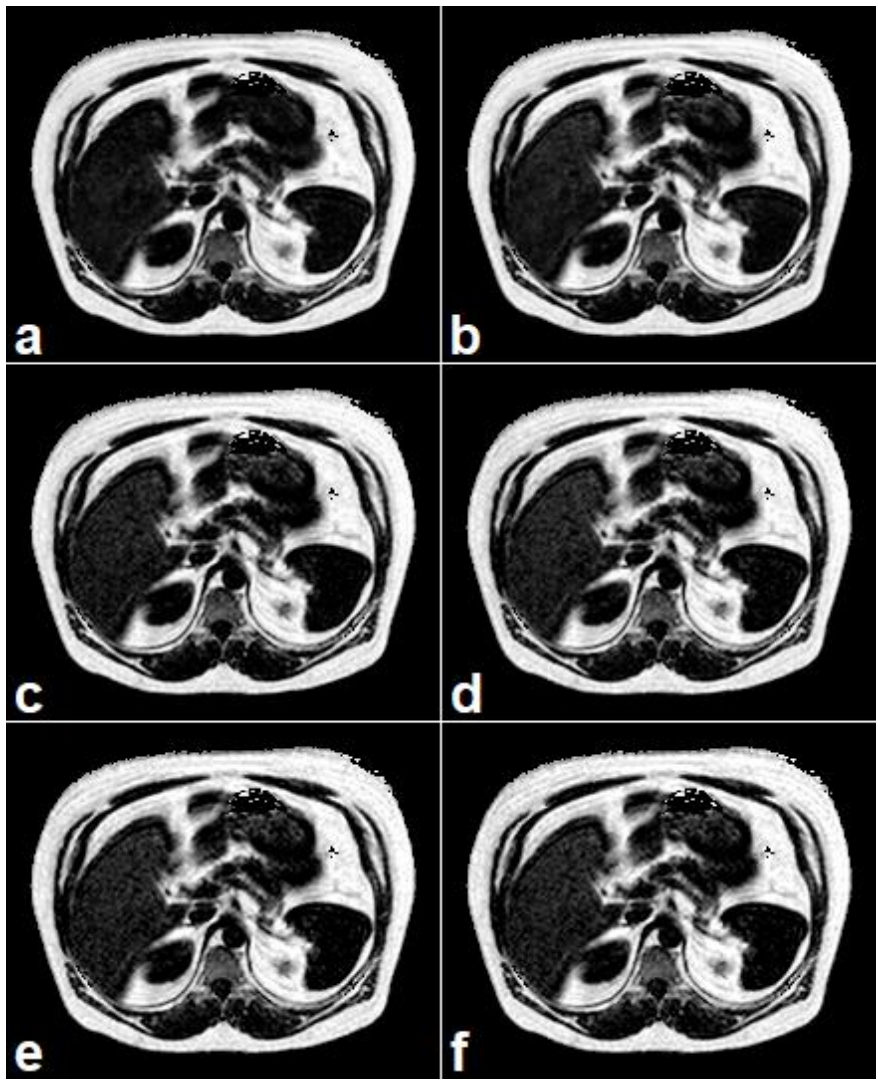

**Figure 4.** Axial fat fraction maps of the abdomen of a subject with a fatty liver (reference fat fraction 12.28%). Background has been removed from all images for clarity. The images are in grayscale with range 0% FF to 100% FF. (**a–e**) Results using neural networks. **a**, Using the 1st echo, **b**, using the 1st and the 2nd echoes, **c**, using the 1st through the 3rd echoes, **d**, using the 1st through the 4th echoes, and **e**, using all 5 echoes. **f**, Reference.

In the supplementary materials there are additional figures. In Supporting Information Figure S1, difference images between the fat fraction maps estimated by the networks in Figure 4 and the reference fat fraction map are shown. In Supporting Information Figure S2, a profile line plot of the slice in Figure 4 is shown. In Supporting Information Figures S3 and S4, water and fat images corresponding to the fat fraction maps inferred by the networks in Figure 4 calculated using Equation 3 and Equation 4, respectively, are shown together with the corresponding reference images. Figures analogous to Figure 4 of different anatomies are also shown. In Supporting Information Figure S5, the upper legs of a subject are shown, in Supporting Information Figure S6, the upper thorax of a subject is shown, and in Supporting Information Figure S7, the knees of a subject with a metal implant in the right knee are shown.

In Figure 5, coronal water signal images of a subject are shown. The image to the left was created using a neural network with the first echo as input, and the image to the right is the reference. The image created using the neural network is very similar to the reference, using more echoes improves the results slightly (not shown because the differences are very slight). However, there are some minor differences between the two images, mainly visible in the intestines and at the interface between the subvolumes, which can be identified by the horizontal strikes. Selected images are representative.

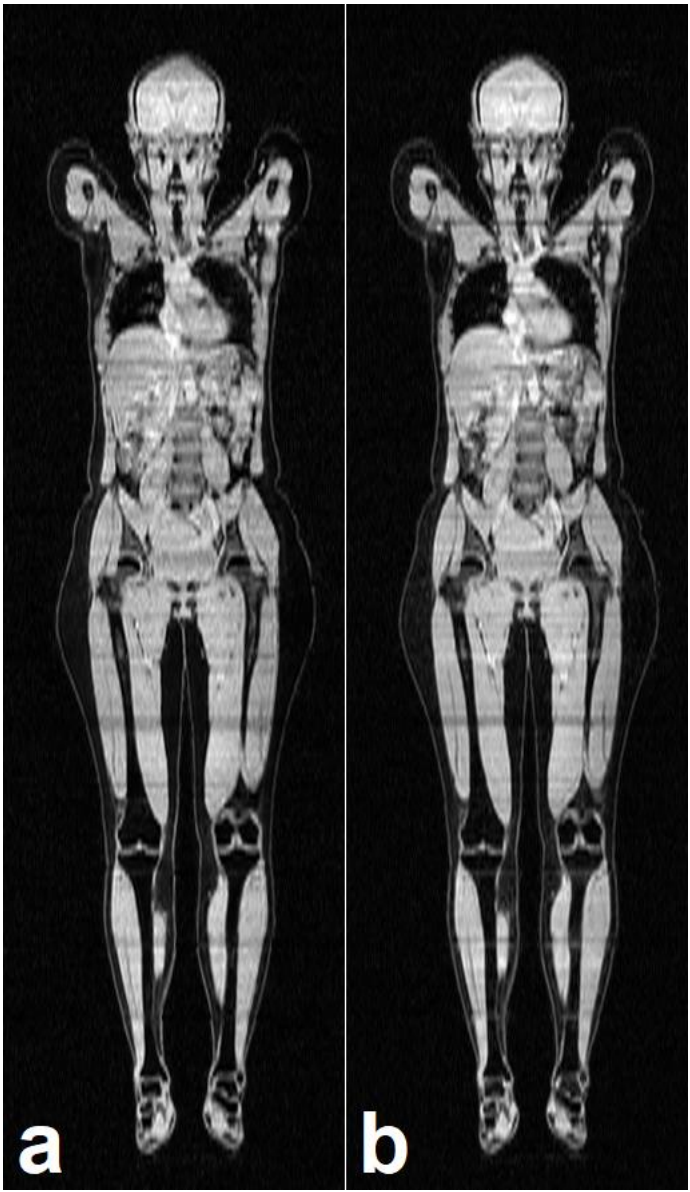

**Figure 5.** Coronal water signal images of a representative subject. **a**, Image created using a neural network with the first echo as input. **b**, Reference.

In general, visual inspection found that the quality of the inferred images was close to that of the references. As exemplified in Figure 4, using only a few echoes often lead to a visibly erroneous liver fat fraction in subjects with an abnormally high liver fat fraction. Other than this, errors were rare. Small but noticeable errors not present in the reference were found in the subcutaneous adipose tissue of 2 subjects with an abnormal amount of subcutaneous adipose tissue. These errors did not completely disappear with more echoes but became less pronounced. Errors near some metal implants were more noticeable compared to the reference when using few echoes, although this discrepancy disappeared when using more echoes. Finally, water–fat swaps were relatively common in the arms in the reference images, probably because of them being in an inhomogeneous area of the main magnetic field, and also present in the anterior subcutaneous adipose tissue of the abdomen in a few subjects, possibly because of motion. These two problems were less pronounced in the images inferred by the neural networks, even when using only 1 echo as input.

### 3.4 Run time

The time taken to train a network was 15.4 h for 1 echo, 17.2 h for 2 echoes, 20.2 h for 3 echoes, 24.8 h for 4 echoes, and 28.6 h for 5 echoes. Using the GPU, inference per slice took 12 ms when using 1 echo, rising to 15 ms when using 5 echoes, corresponding to 3.0 s and 3.7 s per whole-body scan, respectively. When using the CPU, inference per slice took 1.4 s, corresponding to 5.8 min per whole-body scan, regardless of the number of echoes used.

## 4 Discussion

In this study, it has been shown that separation of water–fat signal in whole-body gradient echo images is possible using convolutional neural networks. Separation was possible using only a single echo, even though this is an underdetermined problem, although the results, especially the quantitative measurements, improved when using more echoes, with near identical results to a reference method when provided the same input. The possibility to perform signal separation using only a single echo allows for quicker scanning, which could be useful in situations where a fast scan time is critical.

Only the fat fraction maps were directly inferred using the neural networks, and the water and the fat signals were calculated using these. Tests (not shown) appear to indicate that part of the difference between the calculated water and fat images and the reference is because of this extra calculation step. If the networks had been trained to directly infer the water and the fat signals, the differences with the reference method would likely be reduced.

The networks could have been trained to produce $R_2^*$ and field maps. In the case of $R_2^*$, this was not attempted because the reference $R_2^*$ maps were of very poor quality, presumably because the MRI protocol used was not optimized for this. Field maps were not produced because they very seldom are of interest, but this could be a subject of future study.

It was found that using echoes of both polarities improved the results somewhat, compared to discarding echoes of either polarity. The difference was noteworthy for at least one of the measures in Tables 1–3 when using echoes up to number 4. The difference between using all echoes of both polarities or only all odd echoes was negligible. This is probably because the reference fat fraction maps were calculated using only the odd echoes. Using the 2 first odd echoes resulted in a smaller loss and mean absolute error of the liver fat fraction than using the 2 even echoes. This, despite the even echoes having far more advantageous echo times than the first 2 odd echoes, as can be calculated using Cramér-Rao bounds. Again, this might be because of the odd echoes having been used for calculating the reference fat fraction maps. However, using the 2 first odd echoes resulted in more misclassified livers than using the 2 even echoes, this might, however, have been a coincidence. In contrast, using only the second echo provided better results than using only the first echo.

In this study, versions of the 2D U-net were used. The downsampling steps of the networks allow them to get a greater receptive field, whereas the skip connections allow them to preserve fine details. The receptive field is necessary to prevent water–fat swaps and especially crucial when using only 1 echo as input because otherwise, it would be impossible to find a good solution. Using a 3D architecture would extend the receptive field into an additional dimension, and could potentially improve the results. However, this could lead to an increase in the time needed to perform training.

A GPU is needed to train the networks in a reasonable time, because the training of a single network would likely have taken upward to a month or more using the CPU. Once trained, however, the inference time is reasonable even when using a CPU.

One general drawback of neural networks is that there is no guarantee that they will generalize to other data sets. One hard limitation for the networks used in this study is that the input to any given network is fixed to a certain number of echoes, because this value is hardcoded into the network architecture. Furthermore, input containing data differing too much from that used during training could cause problems (e.g., data collected using a different protocol or type of scanner, subjects belonging to a different age group, or subjects with pathology not seen during training). In this study, errors in the inferred fat fraction maps were observed in the subcutaneous adipose tissue of 2 subjects with an abnormal amount of subcutaneous adipose tissue. This could potentially be because of a lack of this phenotype in the training data. The training data could potentially be expanded by augmentation in an attempt to circumvent this problem. If this was to be attempted, it could be important to make sure that the augmentations are realistic, otherwise the networks might just take longer time to train, without any improvement in performance. Alternatively, a network could be trained with input from multiple studies or use synthetic data to attempt to overcome this problem.

In the current study, the networks were trained with input data from the same scans that were used to create the reference fat fraction maps. A few artifacts were less common in the fat fraction maps inferred by the trained networks than in the reference fat fraction maps. However, any bias present in the reference fat fraction maps will most likely also be presented in the fat fraction maps calculated by the networks. This means that the performance of the networks are limited by the already existing method used to create the reference fat fraction maps and therefore not be very useful. This could be resolved by performing both shorter and longer scans of the same subjects. The echoes from longer scans could then be used to create high quality reference fat fraction maps. A network could then use the echo(es) from the short scans as input and be trained with the high quality reference fat fraction maps as desired output. The inferred fat fraction maps of the network could then be compared to fat fraction maps calculated using a traditional reference method, with the echo(es) of the short scans as input. The results in this study, especially when only using 1 echo, as well as previous studies of similar problems (12–14, 19) suggest that a network may outperform traditional methods. This would need to be tested in a future study, for which several new scans would be required.

It is unclear how the networks with only 1 echo as input manage to calculate the fat fraction. A human observer can often determine the tissue type from magnitude images, and in this way, give a rouge estimate of the fat fraction for each voxel. Furthermore, it is possible to take into account that there are correlations between obesity and fat fraction values. It is possible that the networks use similar approaches. Tests (not presented) showed that it was possible to calculate fat fraction estimates using only the magnitude images as input, although results were less accurate than when using complex images. This indicates that the networks were able to take the phase into account. This could be done, for example, by estimating the contribution to the phase from imperfections in the hardware and then taking the magnetic susceptibility of the different tissues and/or air into account and, in this manner, model the fat fraction.

# 5 Conclusions

It has been shown that it is possible to separate the water and the fat signals of whole-body gradient echo scans using neural networks. Interestingly, separation was possible using only 1 echo, although using more echoes improved the results.

# Acknowledgments

Funding was received from the Swedish Research Council (2016-01040).

Supporting Information

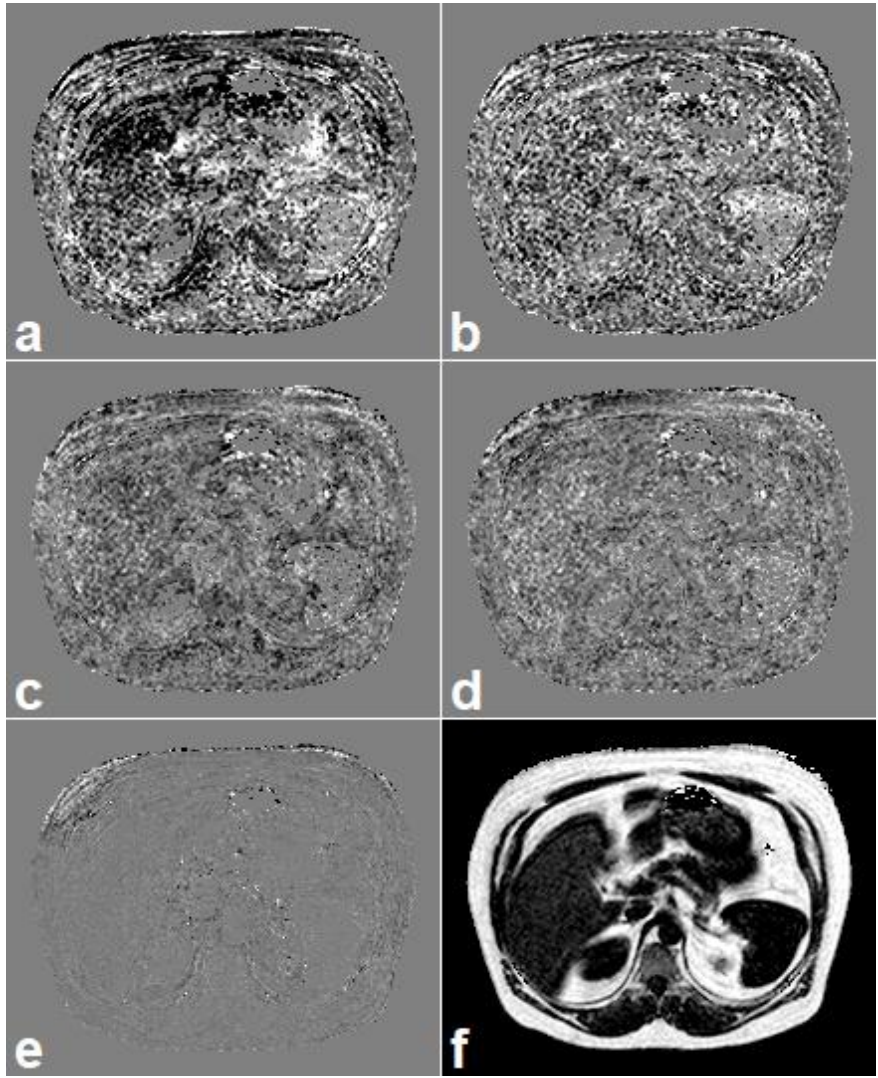

**Figure S1.** Images of the abdomen of a subject with a fatty liver (same subject and slice as in Figure 4). Background has been removed from all images for clarity. (**a–e**) Fat fraction maps inferred using neural networks minus the reference fat fraction map. These images are in grayscale with range −10% FF to 10% FF. **a**, Using the 1st echo, **b**, using the 1st and the 2nd echoes, **c**, using the 1st through the 3rd echoes, **d**, using the 1st through the 4th echoes, **e**, using all 5 echoes. **f**, Reference fat fraction map. This images is in grayscale with range 0% FF to 100% FF.

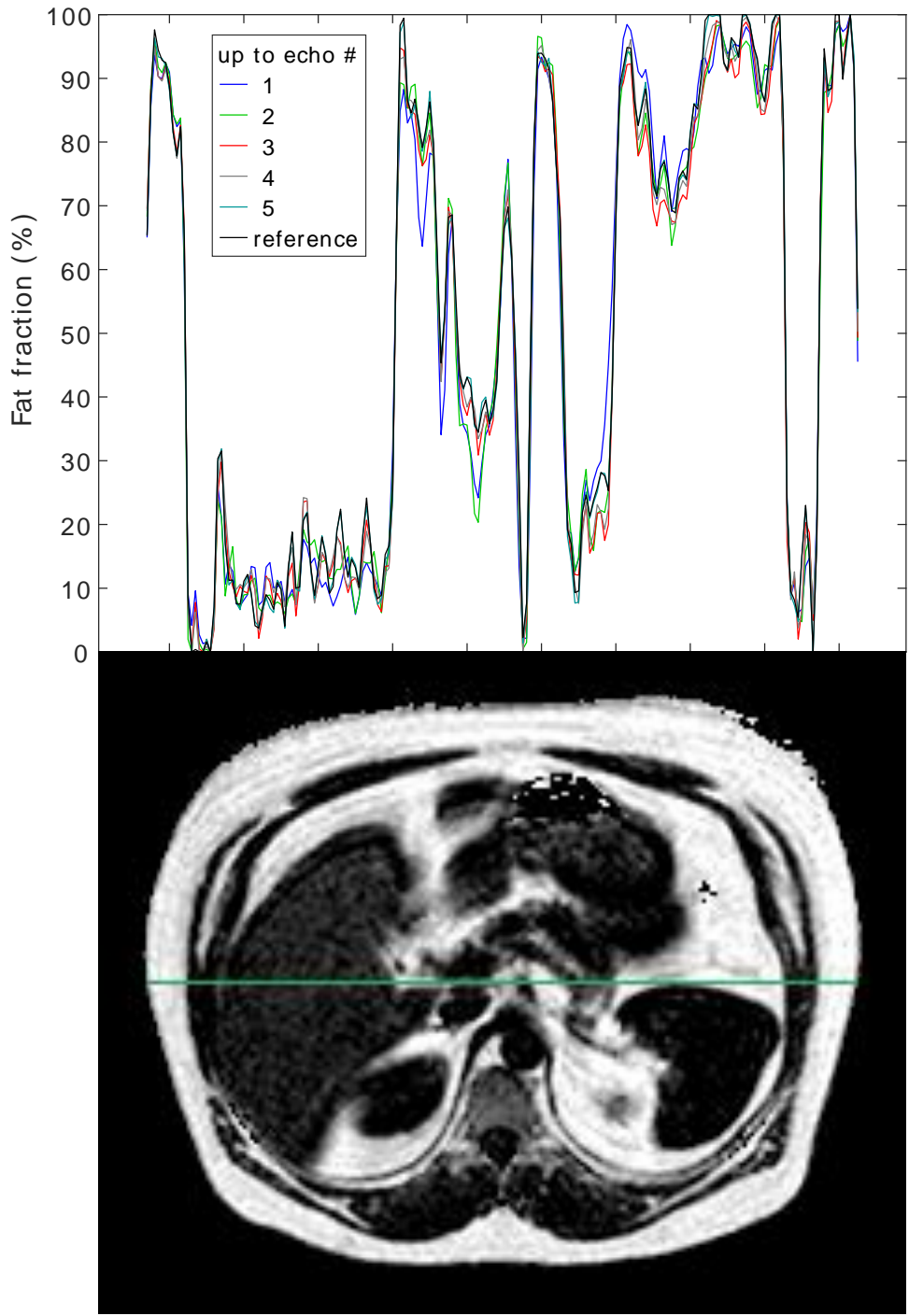

**Figure S2.** Reference fat fraction map of the abdomen of a subject with a fatty liver (same subject and slice as in Figure 4). This images is in grayscale with range 0% FF to 100% FF, and the background has been removed for clarity. The plot shows the fat fractions estimated by the networks using echoes of both polarities and the reference fat fraction along the green profile line.

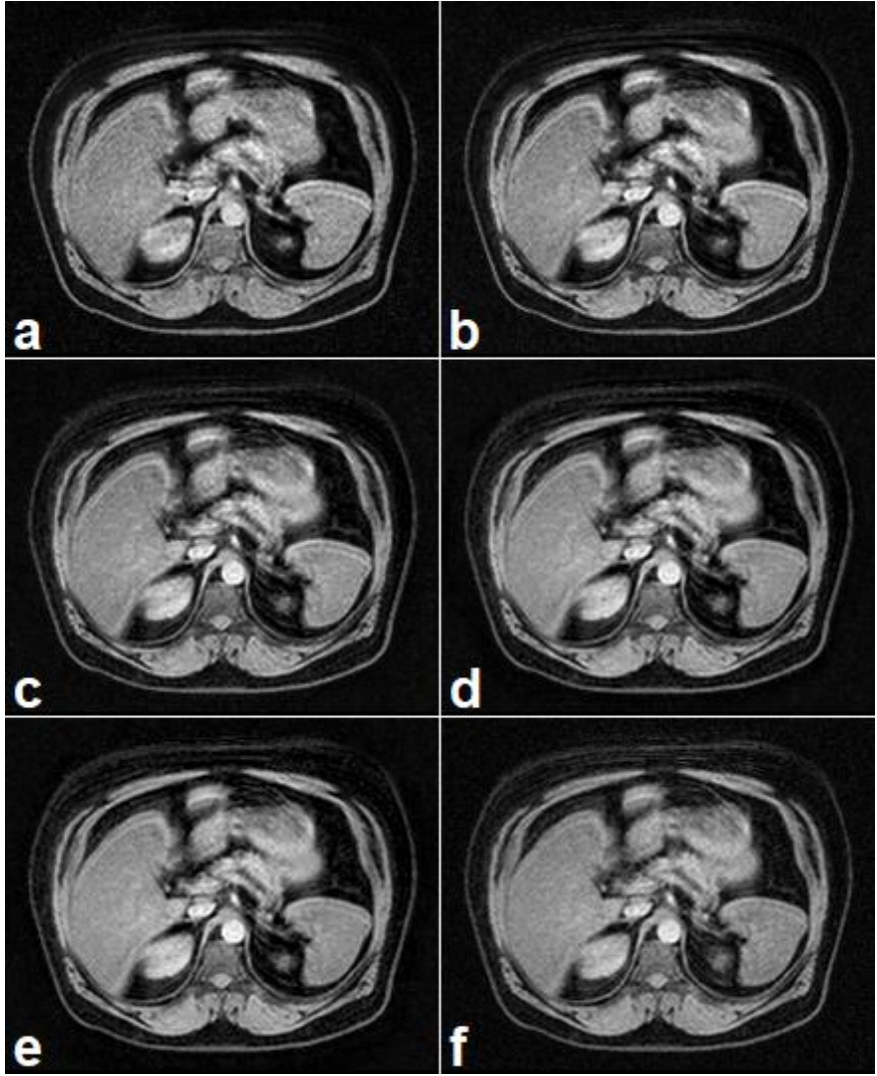

**Figure S3.** Axial water images of the abdomen of a subject with a fatty liver (same subject and slice as in Figure 4). Note that the images in (**a–e**) have been calculated from the inferred fat fraction maps using Equation 3. It is likely that the quality would have been better if the networks had been trained to perform direct inference of water images. (**a–e**) Results using neural networks. **a**, Using the 1st echo, **b**, using the 1st and the 2nd echoes, **c**, using the 1st through the 3rd echoes, **d**, using the 1st through the 4th echoes, **e**, using all 5 echoes. **f**, Reference.

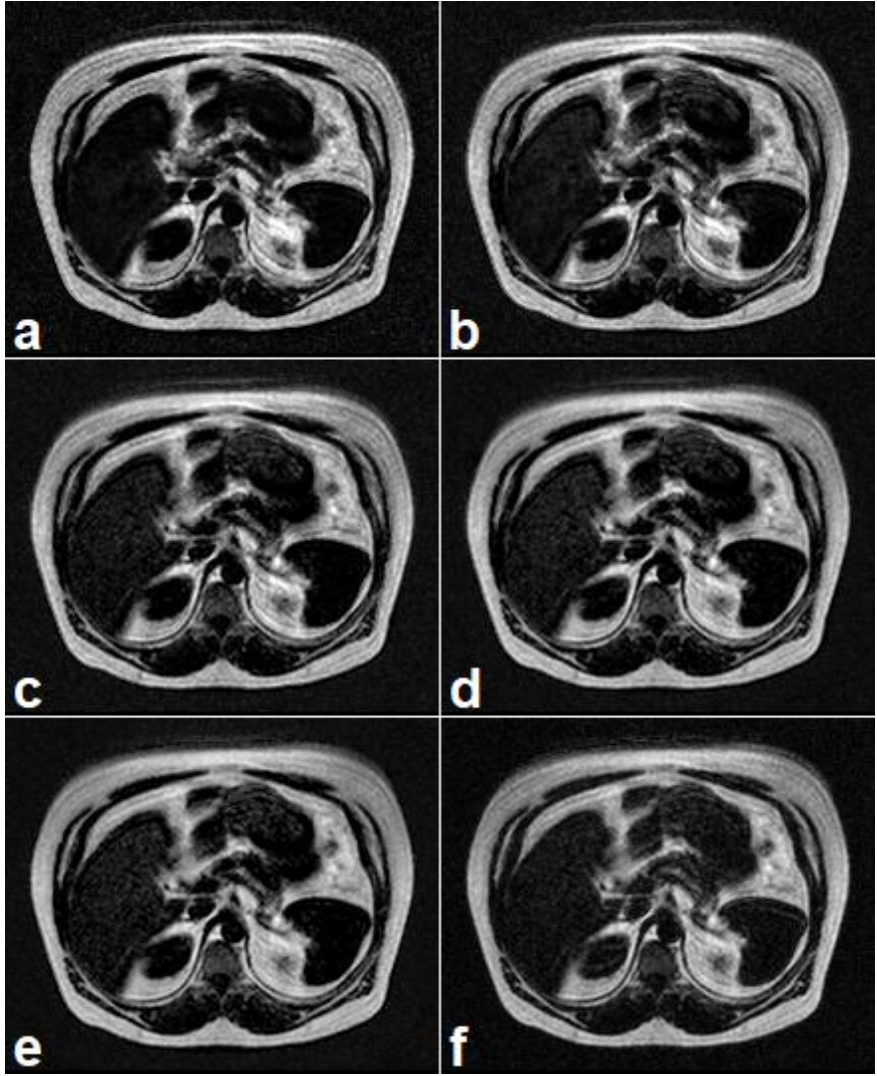

**Figure S4.** Axial fat images of the abdomen of a subject with a fatty liver (same subject and slice as in Figure 4). Note that the images in (**a–e**) have been calculated from the inferred fat fraction maps using Equation 4. It is likely that the quality would have been better if the networks had been trained to perform direct inference of fat images. (**a–e**) Results using neural networks. **a**, Using the 1st echo, **b**, using the 1st and the 2nd echoes, **c**, using the 1st through the 3rd echoes, **d**, using the 1st through the 4th echoes, **e**, using all 5 echoes. **f**, Reference.

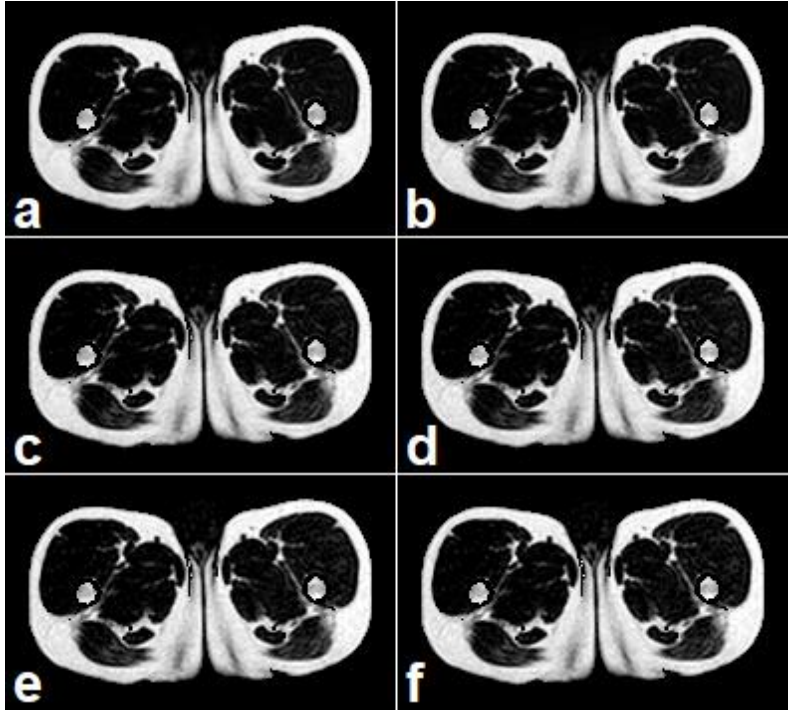

**Figure S5.** Axial fat fraction maps of the upper legs of a subject. Background has been removed from all images for clarity. The images are in grayscale with range 0% FF to 100% FF. (**a–e**) Results using neural networks. **a**, Using the 1st echo, **b**, using the 1st and the 2nd echoes, **c**, using the 1st through the 3rd echoes, **d**, using the 1st through the 4th echoes, **e**, using all 5 echoes. **f**, Reference.

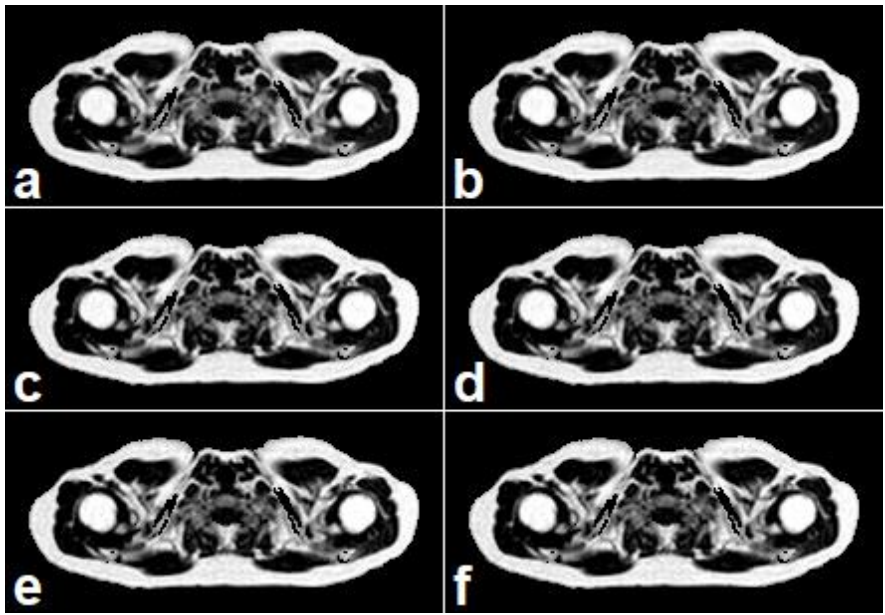

**Figure S6.** Axial fat fraction maps of the upper thorax of a subject. Background has been removed from all images for clarity. The images are in grayscale with range 0% FF to 100% FF. (**a–e**) Results using neural networks. **a**, Using the 1st echo, **b**, using the 1st and the 2nd echoes, **c**, using the 1st through the 3rd echoes, **d**, using the 1st through the 4th echoes, **e**, using all 5 echoes. **f**, Reference.

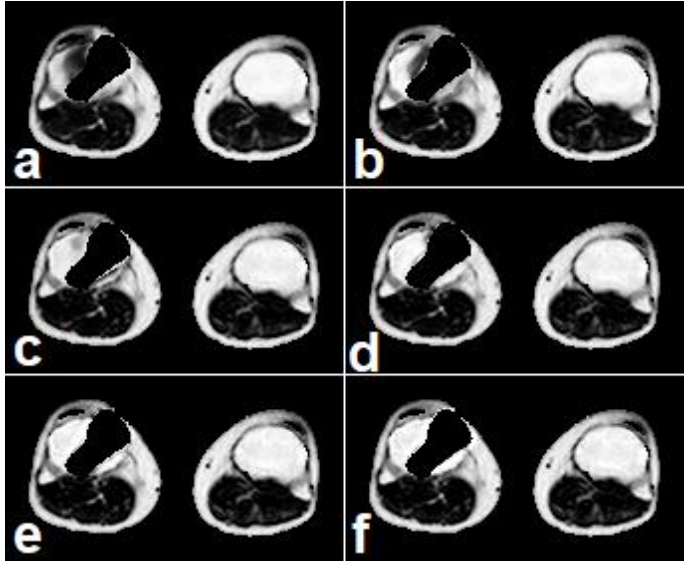

**Figure S7.** Axial fat fraction maps of the knees of a subject with a metal implant in the right knee. This slice was specifically chosen to show the artifact near the metal implant that is visible when reconstructing the FF maps using a few echoes, the artifact is not seen in all slices near the implant. Background has been removed from all images for clarity. The images are in grayscale with range 0% FF to 100% FF. (**a–e**) Results using neural networks. **a**, Using the 1st echo, **b**, using the 1st and the 2nd echoes, **c**, using the 1st through the 3rd echoes, **d**, using the 1st through the 4th echoes, **e**, using all 5 echoes. **f**, Reference.